\documentclass{article}

\usepackage{microtype}
\usepackage{graphicx}
\usepackage{subfigure}
\usepackage{booktabs} % for professional tables
\usepackage{amssymb,amsfonts,amsmath,latexsym,dsfont}
\usepackage{tikz,pgflibraryplotmarks}
\usepackage{graphicx}
\usepackage{mathrsfs}
\usepackage{tcolorbox}

\usepackage{hyperref}

\newcommand{\bfA}{{\bf A}}
\newcommand{\bfB}{{\bf B}}

\newcommand{\bfF}{{\bf F}}

\newcommand{\bfI}{{\bf I}}
\newcommand{\bfJ}{{\bf J}}
\newcommand{\bfK}{{\bf K}}
\newcommand{\bfL}{{\bf L}}

\newcommand{\bfY}{{\bf Y}}

\newcommand{\bftheta}{{\boldsymbol \theta}}

\newtheorem{theorem}{Theorem}

\usepackage[accepted]{icml2019}

\icmltitlerunning{IMEXnet - A Forward Stable Deep Neural Network}

\begin{document}

\twocolumn[
\icmltitle{IMEXnet - A Forward Stable Deep Neural Network}

\icmlsetsymbol{equal}{*}

\begin{icmlauthorlist}
\icmlauthor{Eldad Haber}{equal,eos,xtract}
\icmlauthor{Keegan Lensink}{equal,eos,xtract}
\icmlauthor{Eran Treister}{bgu}
\icmlauthor{Lars Ruthotto}{em}

\end{icmlauthorlist}

\icmlaffiliation{eos}{Department of Earth, Ocean and Atmospheric Sciences, University of British Columbia, Vancouver, Canada}
\icmlaffiliation{xtract}{Xtract AI, Vancouver, Canada}
\icmlaffiliation{em}{Departments of Mathematics and Computer Science, Emory University, Atlanta, GA, USA}
\icmlaffiliation{bgu}{Department of Computer Science, Ben Gurion University of the Negev, Be'er Sheva, Israel}

\icmlcorrespondingauthor{Eldad Haber}{ehaber@eoas.ubc.ca}

\icmlkeywords{Stability, Field of view, Implicit methods,Segmentation, depth of view }

\vskip 0.3in
]

\printAffiliationsAndNotice{\icmlEqualContribution} % otherwise use the standard text.

\begin{abstract}
Deep convolutional neural networks have revolutionized many machine learning and computer vision tasks, however, some remaining key challenges limit their wider use.
 These challenges include improving the network's robustness to perturbations of the input image and the limited ``field of view'' of convolution operators.
 We introduce the IMEXnet that addresses these challenges by adapting semi-implicit methods for partial differential equations.
 Compared to similar explicit networks, such as residual networks, our network is more stable, which has recently shown to reduce the sensitivity to small changes in the input features and improve generalization.
 The addition of an implicit step connects all pixels in each channel of the image and therefore addresses the field of view problem while still being comparable to standard convolutions in terms of the number of parameters and computational complexity.
 We also present a new dataset for semantic segmentation and demonstrate the effectiveness of our architecture using the NYU Depth dataset.
\end{abstract}

\section{Introduction}
\label{intro}

Convolutional Neural Networks (CNN) have revolutionized many machine learning and vision tasks such as image classification, segmentation, denoising and deblurring (see \cite{bengio2009learning,lecun2015deep,Goodfellow-et-al-2016,Hammernik_2017,AVENDI2016108} \mbox{and references within).}

Many different architectures have been proposed and often tailored to specific tasks.
 In recent years, residual networks (ResNets) have shown to be successful in dealing with many different tasks \cite{GomezEtAl2017,he2016identity,he2016deep,LiEtAl2017}.
ResNets have many practical advantages (e.g., ease of training and possible reversibility \cite{YangHuiHe2018,Chang2017Reversible}) and are also supported by mathematical theory due to their link to ordinary differential equations~\cite{weinan2017proposal,NeuralODE2018,HaberRuthotto2017a,CJWE2018,lu2018} and, when dealing with imaging data, partial differential equations~\cite{RuthottoHaber2018}.

The connection between ResNets and differential equations has highlighted the issue of forward stability of the network.
Roughly speaking, a network is forward stable when it does not amplify perturbations of the input features due to, for example, noise or adversarial attacks.
The examples in~\cite{HaberRuthotto2017a,NeuralODE2018} also suggest that stable networks train faster and generalize better.
Keeping a network stable requires attention and can be challenging to control \cite{HaberRuthotto2017a,NAISnet,CJWE2018}.

While it is possible to apply ResNets to many different vision problems, one should differentiate between problems that have a small-dimensional output and problems that have a large-dimensional output.

In a small-dimensional output problem, the image is reduced in dimension to a small vector. For example, in image classification, a small-dimensional vector in $\mathbb{R}^n$ represents the likelihood of the image to be one of $n$ different classes. In this case, the network is being used for dimensionality reduction. For these problems, the image is typically coarsened a number of times before a prediction is made. The coarsening of the image and the use of convolutions allows far-away pixels to communicate, utilizing long-range correlations in the image.

In a large-dimensional output problem, the network generates several different output images and, most commonly, each output image has at least as many pixels as the input image.
An example of this is image segmentation, where each output image represents the probability of each pixel belonging to a certain class.
In depth estimation, image denoising, and image deblurring the output image has the same dimension as the input image, and in many cases contains high spatial frequencies that are absent in the original image.

For these problems, a straightforward extension of the ResNet architecture may not be sufficient because it is unfavorable to coarsen the image and remove high-frequency spatial information. Without coarsening, one is required to work with the original image resolution and compelled to use very deep networks with sufficiently many convolutional layers to model interactions between far away pixels.

This is known as the \emph{field-of-view} problem and has been studied in \cite{NIPS2016_6203} and it is also common in other image processing techniques such as Total Variation image denoising \cite{RudinOsherFatemi92} and anisotropic diffusion \cite{bsmh,weickert}.
Current CNN architectures are limited to either using additional convolutional layers, coarsening, or a combination of both in order to increase their field-of-view. While the receptive field is in general not a problem for small-dimensional output problems such as classification, where coarsening is already being used for dimensionality reduction, it remains a problem for many large-dimensional output problems.
In large-dimensional output problems, image coarsening can be done as a part of the network; however, image interpolation is needed to return to the original image size and provide dense output.
This leads to a different architecture, e.g., the U-net \cite{UNET2015}, which is more complicated than simple ResNets, is less well-understood theoretically, and usually requires many more parameters.

In this paper, we introduce the implicit-explicit network, IMEXnet, and apply it to high-dimensional output problems. Our network is based on simple but effective changes to the popular ResNet architecture and is motivated by semi-implicit techniques for partial differential equations. Such techniques are used for time-dependent problems arising in computational fluid dynamics and imaging when global information is passed within a small number of iterations or time steps \cite{JFNK2011,Schoenlieb2011}.
These techniques address both the stability and the field-of-view issues while adding a negligible number of parameters and computational complexity. We differentiate between large and small-dimensional output problems because the proposed method’s ability to accelerate the propagation of information is most advantageous for high-dimensional output problems.

The paper is structured as follows. In Section~\ref{sec2}, we derive the IMEXnet and explore its theoretical properties.
In Section~\ref{sec3}, we show that our method can be implemented efficiently in existing machine learning packages and demonstrate that it adds only a marginal cost to simple ResNets in terms of the number of operations and required memory.
In Section~\ref{sec4}, we conduct numerical experiments on a synthetic dataset that is constructed to demonstrate the advantages and limitations of the method, as well as on the NYU depth dataset.
We summarize the paper in Section~\ref{sec5}.

\section{Semi-implicit Neural Networks}
\label{sec2}

We first briefly review residual neural networks (ResNets) and outline their limitation in terms of stability and field-of-view problem.
We then derive the basic idea behind our new implicit-explicit IMEXnet as a modification of ResNets.
Finally, we analyze the improved stability of our method and discuss its advantages and disadvantages.

\subsection{Residual Networks}

Our starting point is the $j$-th layer of a ResNet that propagates the features $\bfY_j$ as follows
\begin{eqnarray}
\label{resnet}
\bfY_{j+1} = \bfY_j +h\, f(\bfY_j,\bftheta_j).
\end{eqnarray}
Here, $\bfY_{j+1}$ are the output features of the $j$-th layer, $\bftheta_j$ are the parameters that this layer depends on, and $f$ is a nonlinear function.
In imaging problems, the parameters $\bftheta_j$ typically contain convolution kernels as well as scaling and bias parameters for batch or instance normalization.
Here $h>0$ is a step size that is typically set to $1$.
In particular, we explore the structure proposed in \cite{he2016deep} that has the form
\begin{eqnarray}
\label{layer}
f(\bfY,\bfK_1,\bfK_2,\alpha,\beta) = \bfK_2 \sigma(N_{\alpha,\beta}(\bfK_1\bfY)).
\end{eqnarray}
Here $\bfK_1$ and $\bfK_2$ are built using (typically $3 \times 3$) convolution operators, $N_{\alpha,\beta}(\cdot)$ is a normalization layer that depends on the parameters $\alpha$ and $\beta$, and $\sigma$ is a nonlinear activation function that is applied element-wise.

It is interesting to evaluate the action of this network on some image $\bfY_0$.
At every layer of this network, each pixel communicates with a $5\times 5$ patch around itself.
Therefore, for high-resolution images, many layers are needed in order to propagate information from one side of the image to the other.
This is demonstrated in the top two rows of Figure~\ref{fig1} where two delta functions are propagated through a multi-layer ResNet with 20 layers and $h=1$ and a ResNet with 5 layers and $h=5$.  Comparing the output features it is apparent that the second (4 times less expensive) network is unable to propagate information over large distances.

The above discussion highlights the \emph{field-of-view} problem, i.e.,  that many convolutional layers are needed to model nonlocal interactions between distant pixels. For problems such as image classification, the image is typically coarsened using pooling layers placed between ResNet blocks.
Pooling makes each pixel encompass a larger area and therefore allows information to travel larger distances in the same number of convolution steps.
Coarsening is not applicable in tasks that require a high-dimensional output, as it leads to the loss of important local information.
In these cases, many layers are needed in order to pass information between different parts of the image. This leads to very high computational cost and storage.

At this point, it is worthwhile recalling the differential equation interpretation given to ResNets proposed in \cite{weinan2017proposal,HaberRuthotto2017a}.
In this interpretation the ResNet step \eqref{resnet} is viewed as a forward Euler discretization of the ordinary differential equation (ODE)
\begin{eqnarray}
\label{resnetode}
\dot{\bfY}(t) = f(\bfY(t),\bftheta(t)), \quad \bfY(0)=\bfY_0.
\end{eqnarray}
Here, the features $\bfY(t)$ and weights $\bftheta(t)$ are continuous functions in the (artificial) time that corresponds with the depth of the network.
While it is possible to discretize the system using the forward Euler method (resulting in \eqref{resnet}), many other methods can be used. In particular, in \cite{HaberRuthotto2017a,NeuralODE2018} the midpoint method
was used and Runge Kutta methods were proposed. These methods are all \emph{explicit} methods, i.e., the state, $\bfY$ at time $t_{j+1}$ is explicitly expressed by the states
at previous times.
While such methods enjoy simplicity, they suffer from the field of view problem and a lack of stability.
Indeed, many small steps are needed in order to integrate the ODE for a long time.
In particular, when explicit methods are applied to partial differential equations (PDEs), many time steps are required in order for information to travel on the entire computational mesh.
This problem is well-known and documented in the numerical solution of PDEs, e.g. when solving Navier-Stokes equations \cite{gs}, the solution of flow in porous media \cite{ChenHuanMaBook}, and in cloth simulation for computer graphics \cite{BaraffWitkin1998}.
Hence, the relation of convolutional ResNets to those PDEs described in~\cite{RuthottoHaber2018}, provides an alternative explanation for the field-of-view problem.

\subsection{The Semi-Implicit Network}
One way to accelerate the communication of information across all pixels is to use implicit methods \cite{ap}. Such methods express the state at time $\bfY_{j+1}$ implicitly. For example, the simplest implicit method for ODEs is the backward Euler method where in order to obtain $\bfY_{j+1}$ we solve the nonlinear equation
\begin{eqnarray}
\label{resnetbe}
\bfY_{j+1} - \bfY_j=   h\, f(\bfY_{j+1},\bftheta_{j+1}).
\end{eqnarray}
The backward Euler method is stable for any choice of $h$ when the eigenvalues of the Jacobian of $f$ have no positive real part.
Therefore, it is possible to take arbitrarily large steps in such a network while being robust to small perturbations of the input images due to, for example, noise or adversarial attacks.
Unfortunately, implicit methods can be rather expensive. In particular, the solution of the nonlinear equation \eqref{resnetbe}  is a non-trivial task that can be computationally intensive. 

Rather than using a fully implicit method, we derive a new architecture using the computationally efficient Implicit-Explicit method (IMEX) \cite{ars,arw}.
IMEX is commonly used in fluid dynamics and surface formation and has applied also in the context of image denoising \cite{Schoenlieb2011}.
The key idea of the IMEX method is to divide the right-hand side of the ODE into two parts.
The first (nonlinear) part is treated explicitly and the second (linear) part is treated implicitly.
We design the implicit part so that it can be solved efficiently.
In our context, there is no natural division to an explicit and an implicit part and therefore,
 we rewrite the ODE~\eqref{resnetode} by adding and subtracting a linear invertible matrix
\begin{eqnarray}
\label{resnetode1}
\dot{\bfY}(t) = \underbrace{f(\bfY(t),\bftheta(t)) + \bfL \bfY(t)}_{\rm explicit\ term}\ - \underbrace{\bfL \bfY(t)}_{\rm implicit\ term}.
\end{eqnarray}
Here, $\bfL$ is a matrix that we are free to choose or train. We assume that $\bfL$ is symmetric positive definite matrix that is "easy" to invert. As we show next, we can use a particular $3 \times 3$ convolution model for $\bfL$ that has these properties. Next, we use the forward Euler method for the explicit term and a backward Euler step for the implicit term. The forward propagation through the new network that we call IMEXnet then reads
\begin{eqnarray}
\label{resnetodeImp}
\bfY_{j+1} = (\bfI + h \bfL)^{-1}(\bfY_j + h\bfL \bfY_j + h\, f(\bfY_j,\bftheta_j)),
\end{eqnarray}
where $\bfI$ denotes the identity matrix.

While the forward propagation may appear more complicated than a simple ResNet step, we show below that the computational complexity of the matrix inversion is similar to that of convolution and we emphasize that this construction has some of the favorable properties of an implicit method. Furthermore, we show that through an appropriate choice of the matrix $\bfL$ the network is unconditionally stable. This implies that no exploding modes will occur throughout the network training.  Also, the matrix $(\bfI + h \bfL)^{-1}$ is dense, i.e., it couples all the pixels in each channel of the image in a single step. For problems where the field-of-view is important, such methods can be very effective.

\begin{figure*}
    \includegraphics[width=\textwidth]{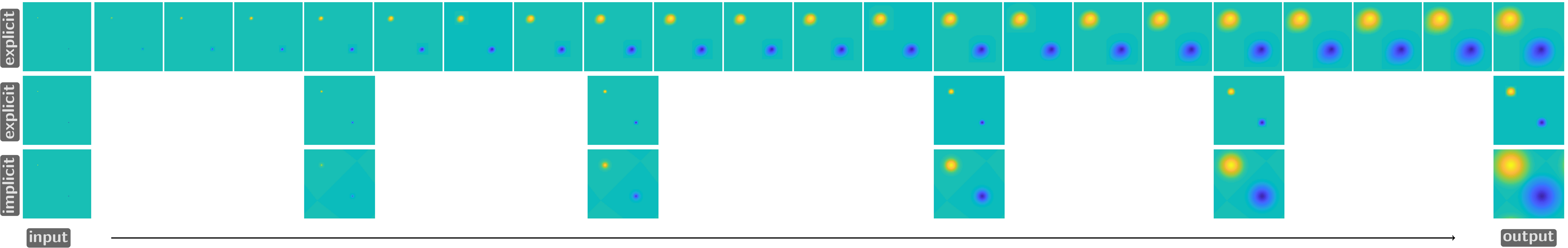}
    \caption{Comparison of explicit and implicit neural networks. Top row: Forward propagation of a test image through an explicit ResNet with 20 time steps. Second row: Explicit ResNet with only five time steps. While the forward propagation is four times faster to evaluate, information is transmitted less effectively. Bottom row: Forward propagation through the proposed semi-implicit network. \label{fig1} }
\end{figure*}

To demonstrate this fact we refer to the third row of Figure~\ref{fig1}.
It shows the forward propagation using the semi-implicit IMEX method where we choose $\bfL$ as a group convolution with the weights
\begin{eqnarray}
\label{lap}
 \bfL = \frac 16 \begin{pmatrix}  -1  &  -4  &  -1 \\ -4  &  20  &  -4 \\ -1  & -4  & -1 \end{pmatrix},
\end{eqnarray}
which is a discrete Laplace operator. We discuss this choice next.
Comparing the output images after only 5 time steps to a ResNet with 20 time steps it is apparent that the IMEX method increases the coupling between far away pixels.

\subsection{Stability of the Method}
\label{sub:stab}
We now discuss the selection of the matrix $\bfL$ and its impact on the stability of the network.
To ensure low computational complexity, we choose to have $\bfL$ as a group convolution.
Assuming, $m$ channels, this implies that the matrix $\bfL$ has the form
\begin{eqnarray*}
\bfL = \begin{pmatrix} \bfL_1 &   &    \\  &  \ddots &  \\
 &      & \bfL_m \end{pmatrix}.
 \end{eqnarray*}
In this way, the implicit step leads to $m$ independent linear systems (which can be parallelized) and it allows us to use tools commonly available in most software packages.
As we show in the next section, such a matrix is stored as a 3D tensor and can be quickly inverted.

We now analyze the IMEXnet in a simplified setting and demonstrate that an adequate choice of $\bfL$ can ensure its stability independent of the step size $h$. 
To this end, we show how to choose $\alpha\geq0$ such that $\bfL = \alpha \bfI$ ensures the stability of the forward propagation for the  model problem
\begin{eqnarray}
\label{modelproblem}
\dot{\bfY}(t) = \lambda \bfY(t), \quad \bfY(t)=\bfY_0.
\end{eqnarray}
Here, $\lambda = -\lambda_{\rm real} + \imath \lambda_{\rm imag}$ with
$0\le \lambda_{\rm real}$ is a given complex number with non-positive real part.  In this case, the norm of the solution $\bfY(t) = \exp(\lambda t) \bfY_0$ is bounded by the norm of $\bfY_0$ for all times $t$.

As discussed above, the usual ResNet is equivalent to the forward Euler method, and for the model problem~\eqref{modelproblem} reads
\begin{equation*}
    \bfY_{j+1} = (1+h\lambda) \bfY_j.
\end{equation*}
This equation is stable (i.e., $\|\bfY_{j+1}\| \leq \|\bfY_j\|$) if and only if
$$ |1+\lambda h| \le 1. $$
Hence, the usual ResNet may be unstable when  $|\lambda|$ is large unless $h$ is chosen small enough, which is computationally expensive.
Now, consider the semi-implicit model with $\bfL=\alpha\bfI$, which can be written as
$$ \bfY_{j+1} = {\frac {1+h\lambda + h\alpha}{1 + h \alpha}} \bfY_j, $$
where a large $h$ can be used as long as $0\le\alpha$ is chosen to ensure the stability of the scheme.
Indeed, since we assume that the real part of $\lambda$ is non-positive, it is always possible to choose $\alpha$ such that $\|\bfY_{j+1}\| \le \|\bfY_j\|$, which implies that stability
is conserved independent of $h$.
The above discussion can be summarized by the following theorem:
\begin{theorem}
Let $\bfJ$ be a given matrix and consider the linear dynamical system $\dot{\bfY}(t) = \bfJ \bfY(t)$.
Assume that the eigenvalues of $\bfJ$ have non-positive real parts, i.e., can be written as $\lambda = -\lambda_{\rm real} + \imath \lambda_{\rm imag}$. Then,  if we choose $\alpha$ such that
\begin{eqnarray}
\label{stb}
{\frac {|\lambda|^2}{2 \lambda_{\rm real}}} - {\frac 1h} \le \alpha, \quad \text{for all }\lambda,
\end{eqnarray}
the magnification factor between layers in the IMEX method \eqref{resnetodeImp} is
$$  \left| {\frac {1+h\lambda + h\alpha}{1 + h \alpha}} \right| \le 1, \quad \text{for all }\lambda,$$
and the method is stable.
\end{theorem}
The proof of this theorem is straight forward by computing the absolute value of the magnification factor.
It is also important to note that as $h\rightarrow 0$ we can choose $\alpha=0$ and keep stability.

Despite its restrictive assumptions, the above theorem provides some intuition about the stability of the forward propagation through the IMEXnet. 
Here, we choose $\bfJ$ as the Jacobian of the layer $f$ in~\eqref{layer} with respect to the input features.
The non-positivity of the real part of the eigenvalues can be ensured by construction (e.g., by setting $\bfK_1 = -\bfK_2^\top$ as in~\cite{RuthottoHaber2018}) and a bound on $|\lambda|$ can be obtained by imposing bound constraints on the convolution weights.
A precise analysis is beyond the scope of this work since the forward problem is nonlinear and non-autonomous.

In our numerical experiments, we pick $\alpha$ to be relatively large (in the range 1-10). We noticed that around this range of values the method is rather insensitive to its choice.

\section{Numerical Implementation and Computational Costs}
\label{sec3}

We show in detail that our network, despite being slightly more complex than the standard ResNet, can be implemented using building blocks that exist in common machine learning frameworks and benefit from GPU acceleration, auto differentiation, etc.
 In particular, we discuss the computation of the implicit step, that is solving the linear system
$$ (\bfI + h\bfL) \bfY = \bfB, $$
where $\bfL$ is constructed above as group-wise convolution and $\bfB$ collects the explicit terms.

To solve the system efficiently, we use the representation of convolution in the Fourier space that states
that the convolution between a kernel $\bfA$ and the features $\bfY$ can be computed as
$$ \bfA * \bfY = \bfF^{-1}((\bfF \bfA) \odot (\bfF \bfY)). $$
where $\bfF$ is the Fourier transform, $*$ is a convolution,  and $\odot$ is the Hadamard element-wise product.
Here, and in the following, we assume periodic boundary conditions on the image data.
This implies that if we need to compute the product of the inverse of the convolution operator defined by $\bfA$ (assuming it is invertible) with a vector, we can simply element-wise divide by the inverse Fourier transform of $\bfA$, i.e.,
$$ \bfA^{-1} * \bfY = \bfF^{-1}( (\bfF \bfY) \oslash (\bfF \bfA)). $$
where $\oslash$ applied element-wise division.

 In our case the kernel $\bfA$ is associated with the matrix
 $$ \bfI + h \bfL, $$
which is invertible, e.g., when we choose $\bfL$ to be positive semi-definite.
Thus, we define
 $$ \bfL = \bfB^{\top} \bfB, $$
 where $\bfB$ is a (trainable) group-wise convolution operator.
 A simple PyTorch code to compute the step is presented in Algorithm \ref{alg1}.

\begin{algorithm*}
   \caption{pyTorch implementation of the implicit convolution.}
   \label{alg1}
\begin{tcolorbox}
\scriptsize
\begin{verbatim}
def diagImpConv(x, B,h):
    n = x.shape
    m = B.shape
    mid1 = (m[2] - 1) // 2
    mid2 = (m[3] - 1) // 2
    Bp = torch.zeros(m[0],n[2], n[3],device=B.device)
    Bp[:, 0:mid1 + 1, 0:mid2 + 1] = B[:, 0, mid1:, mid2:]
    Bp[:, -mid1:, 0:mid2 + 1]     = B[:, 0, 0:mid1, -(mid2 + 1):]
    Bp[:, 0:mid1 + 1, -mid2:]     = B[:, 0, -(mid1 + 1):, 0:mid2]
    Bp[:, -mid1:, -mid2:]         = B[:, 0, 0:mid1, 0:mid2]

    xh  = torch.rfft(x,  2, onesided=True)
    Bh  = torch.rfft(Bp, 2, onesided=True)
    t   = 1.0/(1.0 + h * (Bh[:, :, :, 0] ** 2 + Bh[:, :, :, 1] ** 2) )
    xBh = torch.zeros(n[0], n[1], n[2], (n[3] + 2) // 2, 2,device=B.device)
    for i in range(n[0]):
        xBh[i, :,:, :, 0] = xh[i, :, :, :, 0]*t
        xBh[i, :,:, :, 1] = xh[i, :, :, :, 1]*t
    xB = torch.irfft(xBh, 2, onesided=True,signal_sizes = x.shape[2:])
    return xB
\end{verbatim}
\end{tcolorbox}
\end{algorithm*}

 Using Fourier methods we need to have the convolution kernel at the same size as the image we convolve it with.
 This is done by generating an array of zeros that has the same size of the image and inserting the entries of the convolution into the appropriate places.
 The techniques is explained in detail in~\cite{nagyHansenBook}.

Let us now discuss the memory and computational effort involved with the method.
To be more specific, we analyze the method for a single ResNet layer with $m$ channels, applied to an image of size $s\times s$. Assuming that the stencil size of the convolutions is ${\cal O}(1)$, applying such a layer to an image requires a computational cost of ${\cal O}(m^2 s^2)$
 operations
and a memory that is ${\cal O}(m^2)$ to store the weights.

For the implicit networks, we have the usual explicit step followed by an  implicit step. The implicit step is a group convolution is requires ${\cal O}(m (s \log(s))^2)$ additional operations, where the $s \log(s)$ term results from the Fourier transform. Since $\log(s)$ is typically much smaller than $n$ the additional cost of the implicit step is insignificant.

The memory footprint of the implicit step is also very small. It requires only ${\cal O}(m)$ additional coefficients. For problems where the number of channels is larger than say, $100$ this cost represents less than $1\%$ additional storage.
Thus, the improvement we obtain to ResNet comes with a very low cost of both computations and memory.

\section{Numerical Experiments}
\label{sec4}

In this section, we conduct two numerical experiments that demonstrate the points discussed above. In the first problem, we experiment with semantic segmentation of a synthetic dataset that we call the Q-tips dataset.
We designed this dataset to expose the limitations of explicit methods and demonstrate the improvements of semi-implicit methods.
In the second example, we show the advantages of our approach on the NYU Depth Dataset V2 that contains images of different room types together with their depth. The goal of the training here is to predict the depth map given the image of the room. While the two problems are different in their output they share the need for nonlocal coupling across large distances in order to deliver an accurate prediction.

\subsection{The Q-tips Dataset}

We introduce a synthetic semantic segmentation dataset intended to quantify the effect of a network's receptive field. 
In this dataset, every image contains a single object composed of a rectangular gray midsection with either a white or black square at each end. We define the object classes according to the combination of markers present, resulting in three classes (white-white, white-black, and black-black). The dataset is specifically designed to require a receptive field that encompasses the entire object for accurate classification. If information from both ends of the object is not available to a pixel in the output, then the problem is ambiguous.

For the following experiments we generate a dataset of 1024 training examples and 64 validation examples. Each $64\times 64$ image consists of a single object of length, $l$, width, $w$, and orientation, $r$, randomly selected from a discrete uniform distribution, where $l \in \mathcal{U}\{32, 60\}$, $w \in \mathcal{U}\{4, 8\}$, and $r \in \mathcal{U}\{-180, 180\}$.

In order to evaluate the effect of our proposed semi-implicit architecture, we train two nearly identical 12-layer IMEXnet with weights that are randomly initialized from a uniform distribution on the interval $[0,1)$. The opening layer expands the single channel input to 64 channels, and the width is subsequently doubled every 4 layers to result in a 224 channel output before the classifier. Neither network contains any pooling layers, and the convolution layers are padded such that the input and output are of the same resolution. In order to make one of the networks purely explicit, we set $\mathbf{L}=\mathbf{0}$, which will prevent any implicit coupling, effectively resulting in a ResNet. In both cases, we use stochastic gradient descent to minimize the weighted cross entropy loss for 200 epochs with a learning rate of 0.001 and a batch size of 8. The loss is weighted according to the normalized class frequencies calculated from the entire dataset in order to address the class imbalance due to the background.

For comparison we present various error measurements for both networks on the validation dataset in Table~\ref{tab:results-table}.
\begin{table}[t]
\begin{center}
\begin{small}
\begin{sc}
\begin{tabular}{l|c|c|c|r}
\toprule
Network & Parameters & IOU & Loss & Accuracy \\
\midrule
IMEXnet   & 2701440 & 0.926 & 0.0982 & 99.56 \\
ResNet    & 2691648 & 0.741 & 0.3332 & 98.18 \\
\bottomrule
\end{tabular}
\end{sc}
\end{small}
\caption{Comparison of semi-implicit IMEXnet and explicit ResNet on the synthetic Q-tips validation set.
\label{tab:results-table}}
\end{center}
\vskip -0.1in
\end{table}
Note that the IMEX method did much better in terms of loss and intersection over unions (IOU). The pixel accuracy counts the background and therefore is somewhat misleading.
An example of the results on the testing set is plotted in Figure~\ref{fig:results}.
\begin{figure*}
\centering
\label{fig:results}
\newcommand{\image}[1]{\includegraphics[width=.13\linewidth,trim=102 35 100 38, clip=true]{#1}}
\begin{tabular}{@{}cccc@{}}
    Image & Segmentation & IMEX Predicition & ResNet Predicition \\
  \image{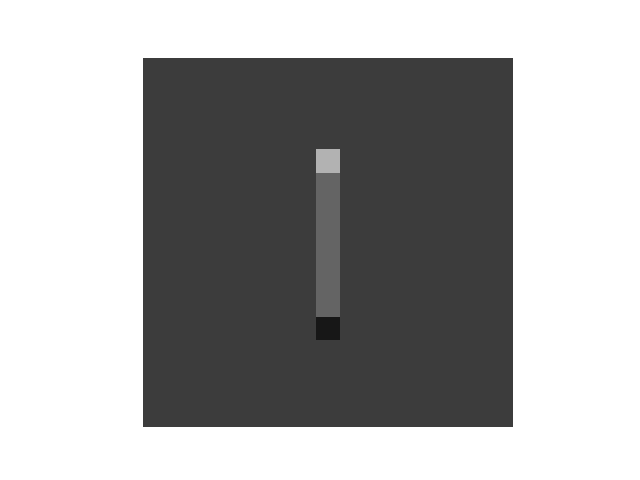} &
  \image{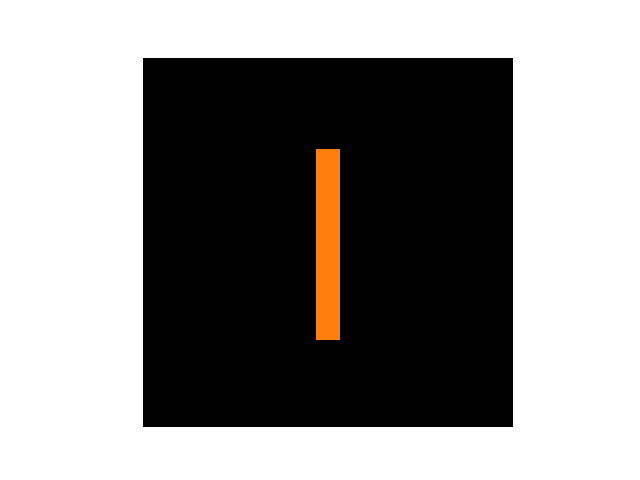} &
  \image{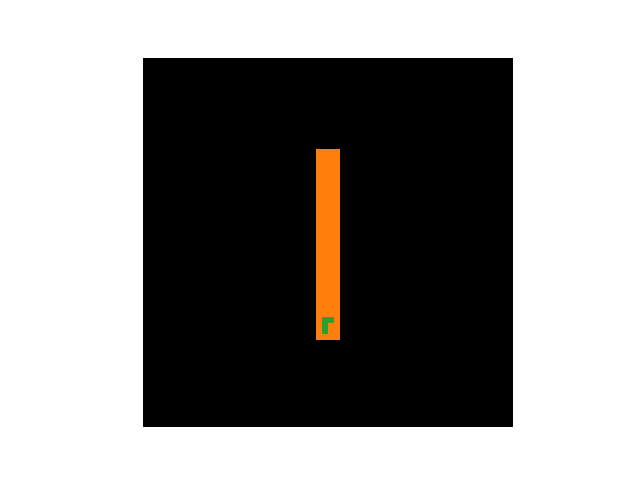} &
  \image{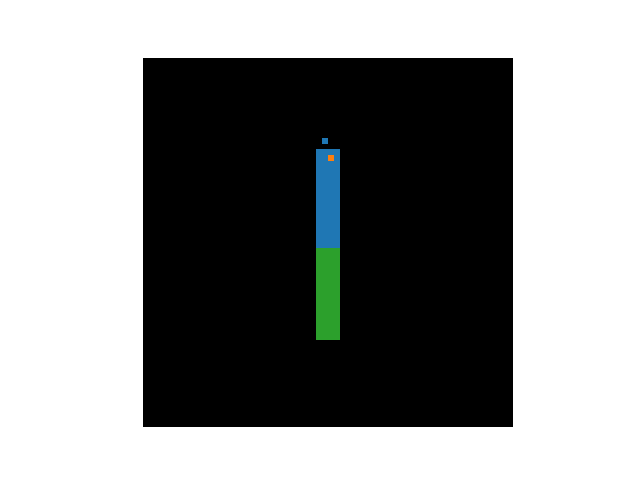} \\

  \image{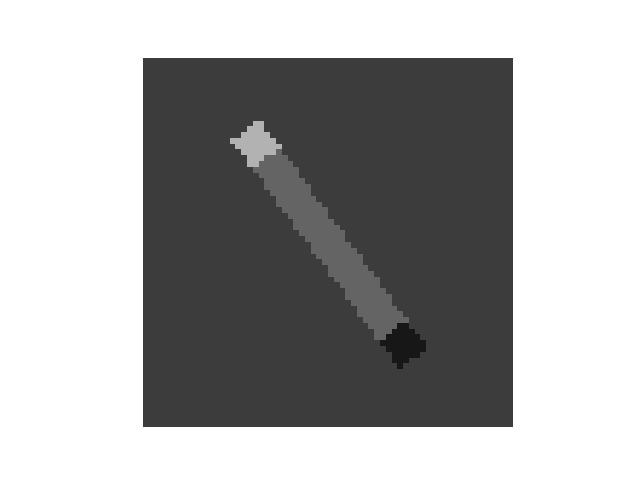} &
  \image{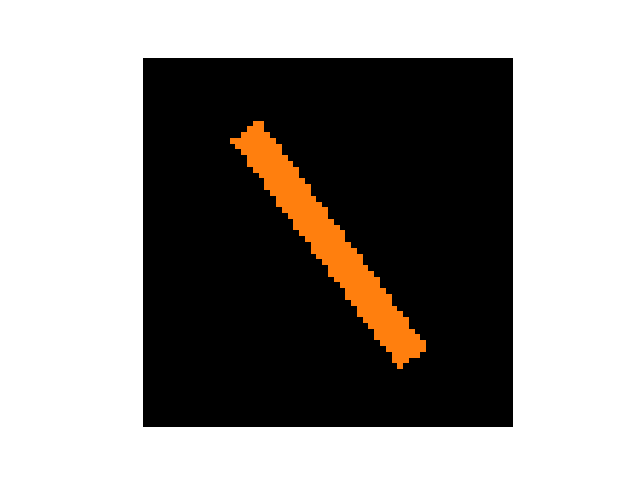} &
  \image{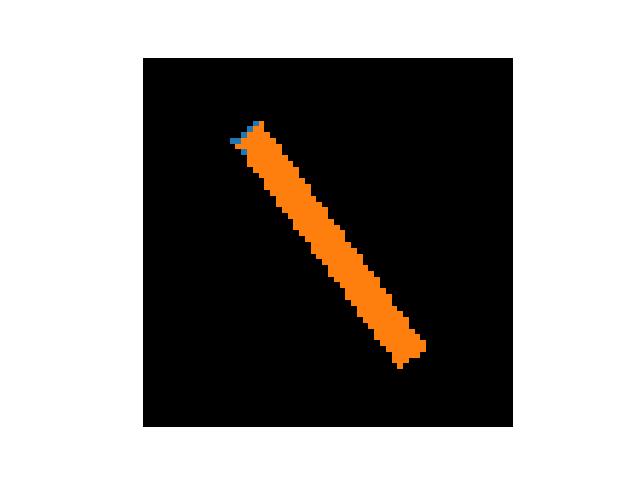} &
  \image{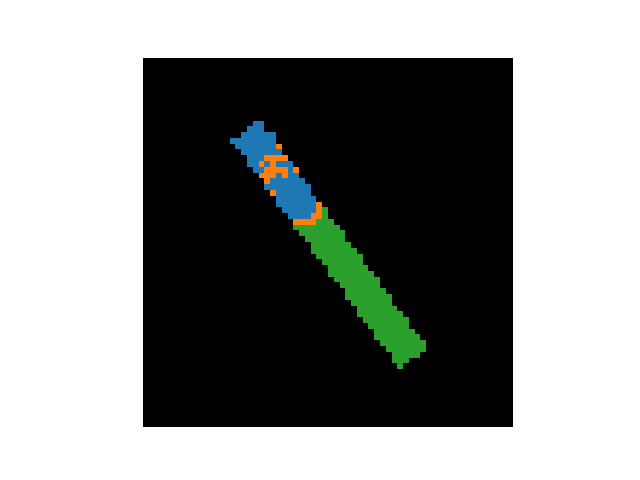} \\

  \image{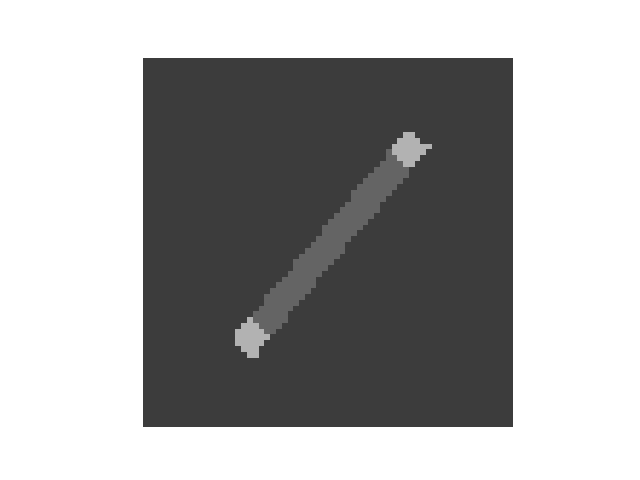} &
  \image{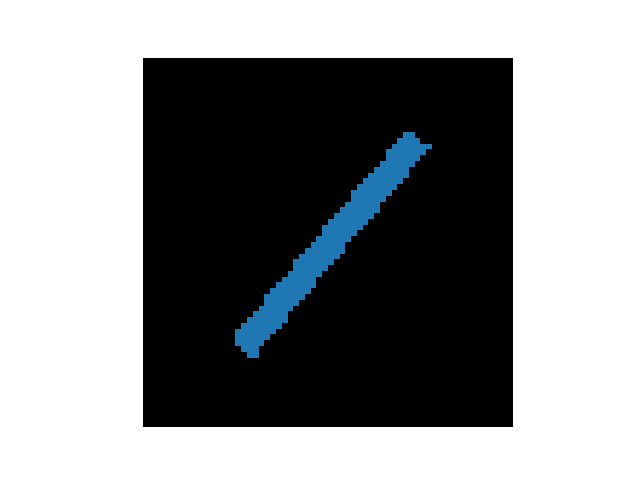} &
  \image{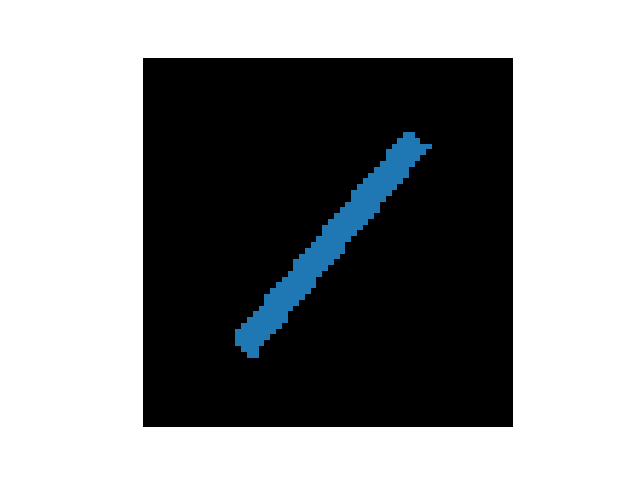} &
  \image{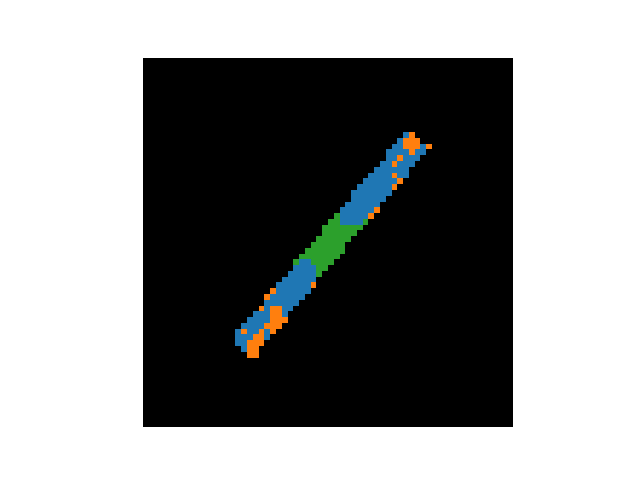} \\

  \image{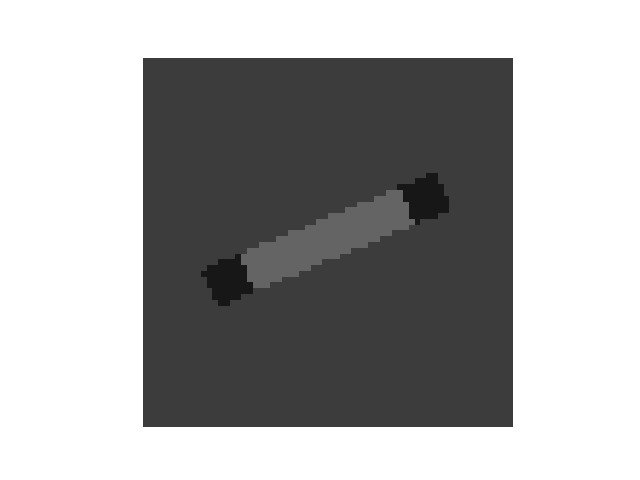} &
  \image{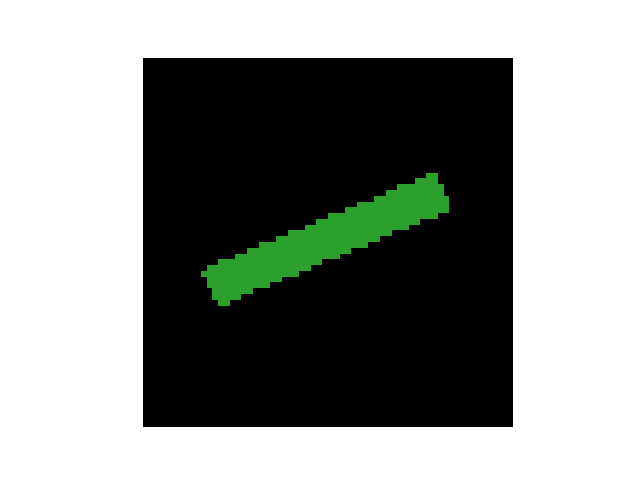} &
  \image{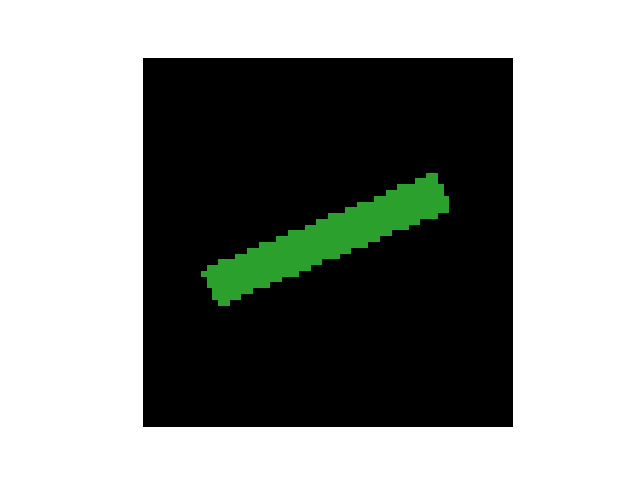} &
  \image{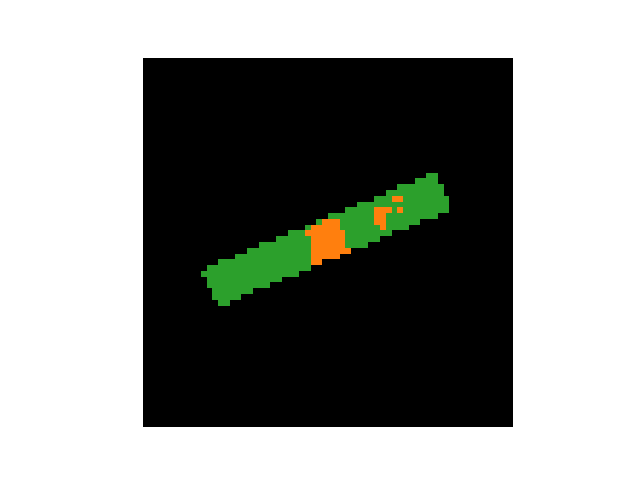} \\

\end{tabular}
\caption{Segmentation results on the Q-tips dataset. We present the image (first column), the ground truth segmentation (second column), our method's prediction (third column), and the ResNet's prediction (last column).}
\label{fig:results}
\end{figure*}

The table and images demonstrate that the ResNet cannot label the image correctly. The centerpiece of the rod is not continuously segmented with the end of the rod. In particular, the center of the rod is randomly classified as one of three object classes. Adding an implicit layer with a negligible memory footprint and computational complexity manages to resolve the problem, obtaining a near perfect segmentation.

\subsection{The NYU Depth Dataset}

\begin{figure}
\begin{tabular}{cc}
\includegraphics[width=3.7cm]{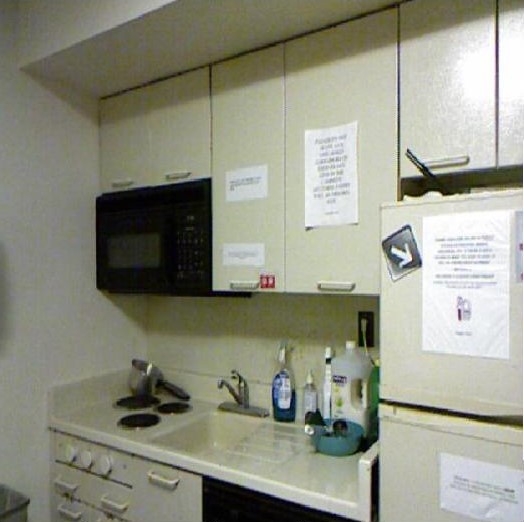} &
\includegraphics[width=3.7cm]{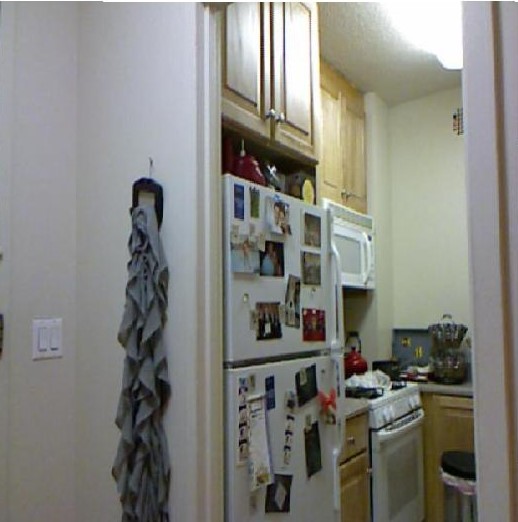} \\
\includegraphics[width=3.7cm]{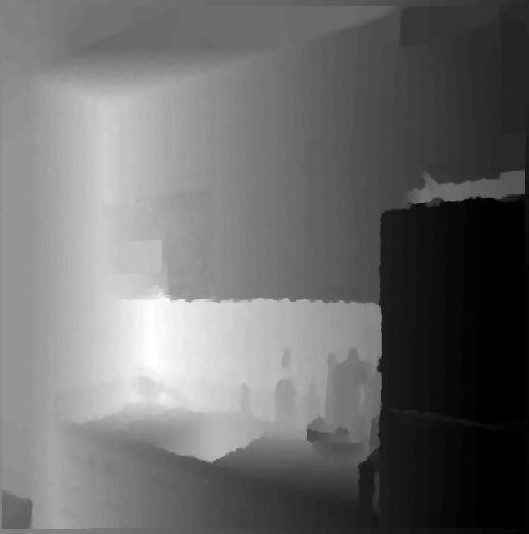} &
\includegraphics[width=3.7cm]{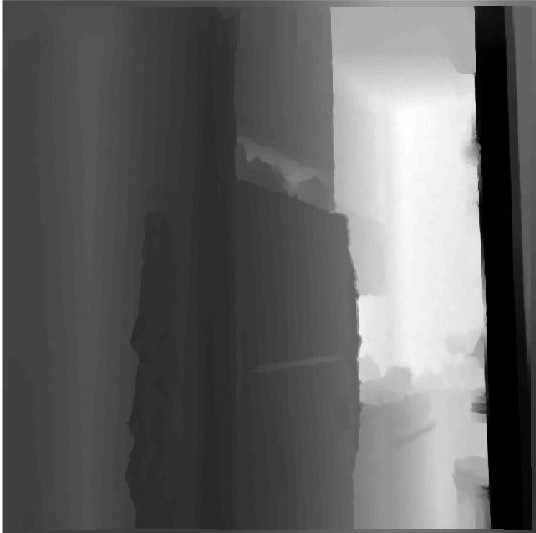} \\
\end{tabular}
\caption{Examples from the NYU depth dataset. Two sets of kitchens are used together with their depth maps.\label{figNYU}}
\end{figure}

\begin{figure}
\begin{center}
\includegraphics[width=2.5cm]{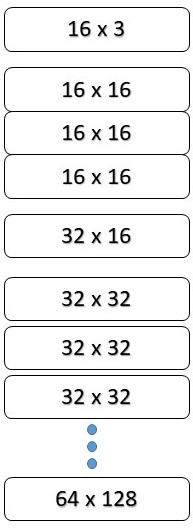}
\caption{The ResNet used for the NYU dataset. \label{resnetfig}}
\end{center}
\end{figure}

The NYU-Depth V2 dataset is a set of  indoor images recorded by both RGB and Depth cameras from the Microsoft Kinect. Four different scenes from the dataset are plotted in Figure~\ref{figNYU}.
The goal of our network is to use the RGB images in order to predict the depth images. We use a subset of the dataset, made of the kitchen scene in order to train a network to achieve this task.
The network contains three ResNet blocks and three bottle-neck blocks and is plotted in Figure~\ref{resnetfig}.

\begin{figure}
    \begin{center}
\includegraphics[width=6.5cm]{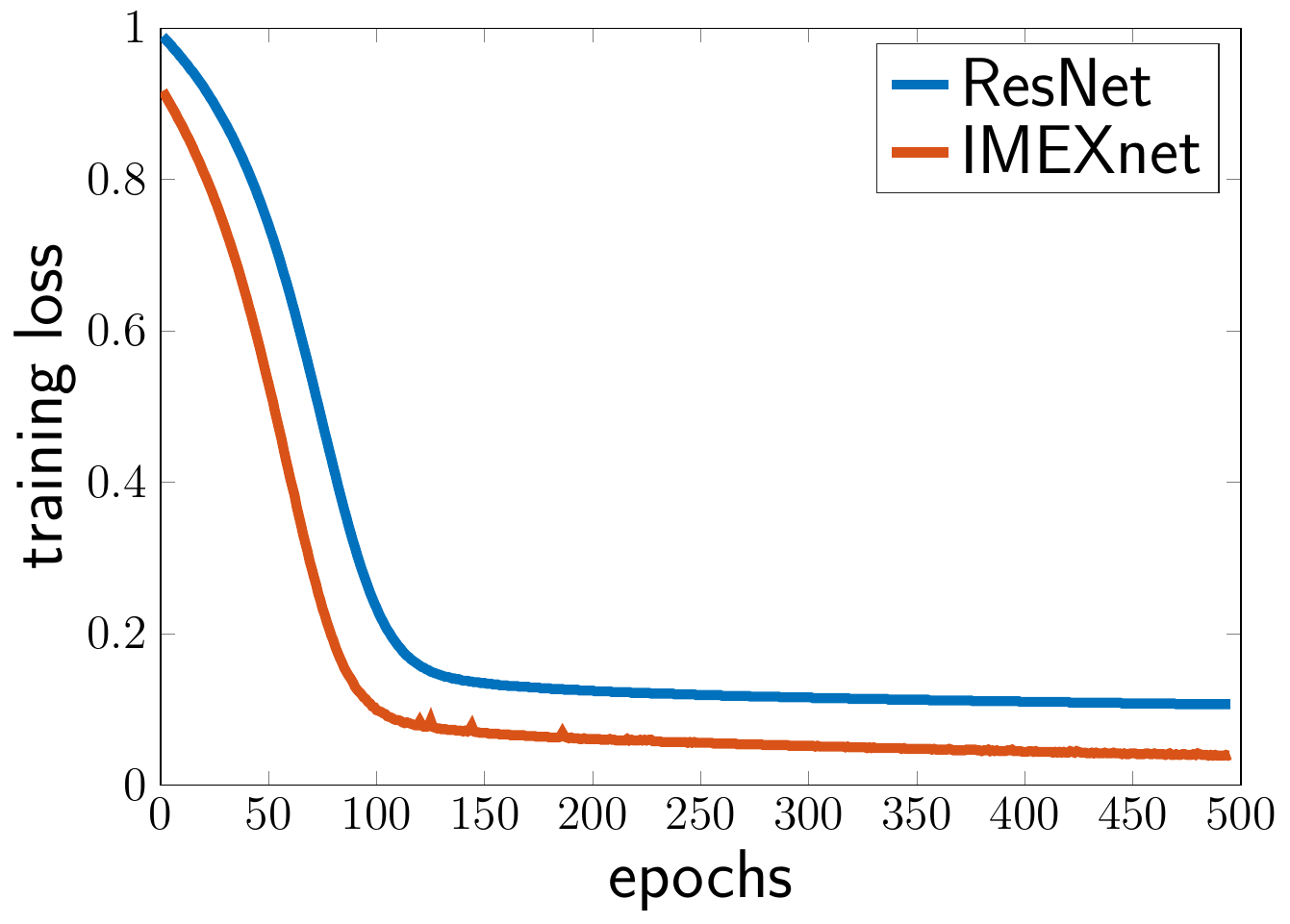}        
    \end{center}
\caption{Convergence of the training using ResNet and IMEX methods on the NYU depth dataset. \label{conv}}
\end{figure}

The ResNet has only 506,928 parameters. For the IMEXnet we use an identical network but add implicit layers. This adds only roughly 27,000 more parameters
for the implicit network. We use 500 epochs to fit the data. The initial $L_2$ misfit
is $8.2$. Using the ResNet architecture we are able to decrease the misfit to $1.10 \times 10^{-2}$. The small addition of parameters for the implicit method allows us to fit the data to $2.9 \times 10^{-3}$.
The convergence of the two methods is presented in Figure~\ref{conv}.

\begin{figure}[h]
\begin{tabular}{cc}
\includegraphics[width=3.5cm]{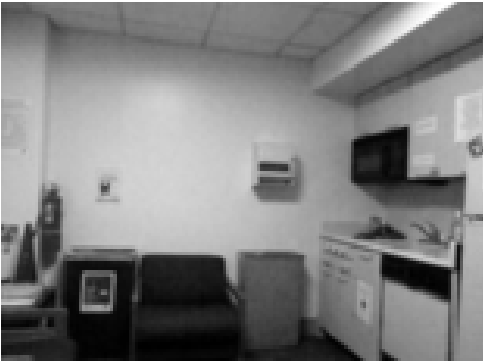} &
\includegraphics[width=3.5cm]{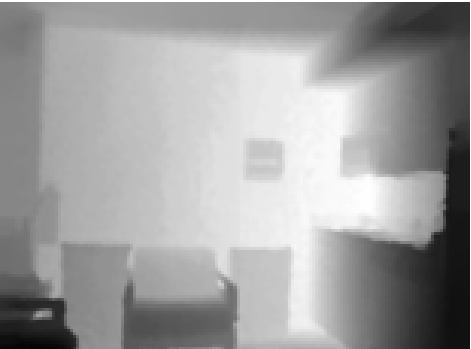} \\
Kitchen scene &  Depth map \\
\includegraphics[width=3.5cm]{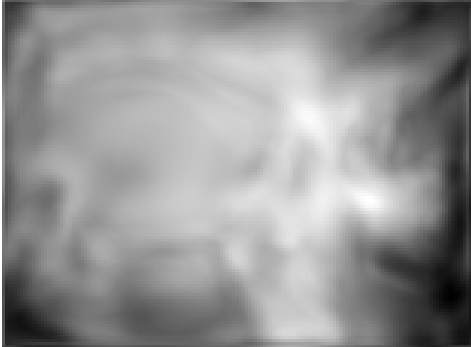}  &
\includegraphics[width=3.5cm]{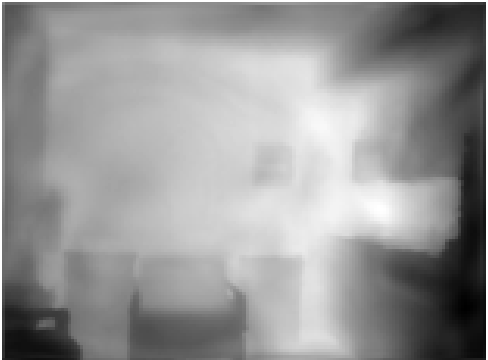}   \\
ResNet recovery & Implicit net recovery
\end{tabular}
\caption{The testing image from the NYU dataset.  \label{depthDS}}
\end{figure}

We found that for learning one set of images (i.e. kitchens), it is possible to use rather few examples. For the kitchen dataset we used only $8$ training images, $2$ validation image and one test image. The results on the test image is plotted in Figure~\ref{depthDS}.
We observe that we are able to obtain good results even when the number of images used for the training is small. These results echo the results obtained for image filtering using variational networks presented in \cite{PockDepth}.

Comparing the ResNet and the IMEXnet we see that the predictions of the IMEXnet is smoother than the one obtained from the ResNet. This is not surprising as the implicit step in the IMEXnet can be seen as a smoothing step. We also note that in our numerical experiments we have found that  the implicit method is less sensitive to initialization. This is not surprising as the implicit step adds stability. Indeed, if we compare an explicit ResNet with an implicit one with the same kernels, the analysis suggests that while the explicit network may be unstable, the implicit one is stable, assuming $\bfL$ is chosen appropriately.

\section{Summary}
\label{sec5}

In this paper, we introduce the IMEXnet architecture for computer vision tasks that is inspired by semi-implicit IMEX methods commonly used to solve partial differential equations.
Our new network extends standard ResNet architectures by adding implicit layers that involve a group-wise inverse convolution operator after each explicit layer.
We have discussed and exemplified that this approach can resolve the field of view problem as well as the issue of the forward stability of the network.
This makes this type of network suitable for problems where the dimension of the output is similar to the dimension of the input, such as semantic segmentation and depth estimation, and where nonlocal interactions are needed. 

We exemplify, using PyTorch, that our method can be implemented efficiently using the available built-in functions.
The computational complexity and memory allocation that is added by using the implicit step is small compared with the complexity and memory needed by the ResNet. We have shown that although the method has marginally larger cost compared with ResNet, it can be much more efficient in training as well as in validation on two simple model problems of semantic segmentation and depth estimation from images. Our code is available at \url{https://github.com/HaberGroup/SemiImplicitDNNs}.

While we have explored one semi-implicit method \eqref{resnetode1}, it is important to realize that we are free to choose other models with similar properties. One attractive option is to remove the matrix $\bfL$ from the explicit part, and consider the diffusion-reaction problem
\begin{eqnarray}
\label{resnetode2}
\dot{\bfY}(t) = f(\bfY(t),\bftheta(t)) - \bfL \bfY(t).
\end{eqnarray}
These types of equations have been used extensively to model nonlinear phenomena such as pattern formation and can have interesting behavior, e.g., it can lead to nonlinear waves. These systems were already studied by Turing \cite{Turing52}
and have been studied extensively in \cite{murray2,Ruuth2,Witkin1991}.
A similar treatment using an IMEX integration scheme leads to 
\begin{eqnarray}
\label{drResnet}
\bfY_{j+1} = (\bfI + h \bfL)^{-1}(\bfI  + h\, f(\bfY_j,\bftheta_j)).
\end{eqnarray}
Performing a similar analysis than the one in Section~\ref{sub:stab}, we see that this network has better stability properties. In particular, we can remove the restriction on having the real parts of the eigenvalues of $\bfJ = \nabla_{\bfY}f^\top$ to be positive. Since these types of equations are used for pattern formation, this type of network may have advantages when considering an output which has patterns, e.g. segmentation of texture. Detailed experimentation and evaluation of this approach is an item of future work.

Since semi-implicit methods are essential in many fields, we believe that this paper illustrates that they can also play a large role in the field of machine learning using deep nets.

\section*{Acknowledgements}
EH's work is supported by the Natural Sciences and Engineering Research Council of Canada (NSERC).
KL's work is supported by the Mitacs Accelerate program and Xtract AI.
LR’s work is supported by the US National Science Foundation (NSF) awards DMS 1522599 and  DMS 1751636

\bibliographystyle{icml2019}
\bibliography{biblio.bib}

\end{document}